\documentclass[a4paper]{article}

\usepackage{ISCSLP_v2}
\usepackage{multirow}
\usepackage{dashrule}

\title{Improving Gated Recurrent Unit Based Acoustic Modeling \\ with Batch Normalization and Enlarged Context}
\name{Jie Li$^1$, Yahui Shan$^2$, Xiaorui Wang$^1$, Yan Li$^1$}
\address{
  $^1$Kwai, Beijing, P.R. China \\
  $^2$School of Information and Electronics, Beijing Institute of Technology, Beijing, P.R.China}
\email{\{lijie03, wangxiaorui, liyan\}@kuaishou.com, 2120160735@bit.edu.cn}

\begin{document}

\maketitle
\begin{abstract}
The use of future contextual information is typically shown to be helpful for acoustic modeling. 
Recently, we proposed a RNN model called minimal gated recurrent unit with input projection (mGRUIP), in which a context module namely \emph{temporal convolution}, is specifically designed to model the future context. This model, mGRUIP with context module (mGRUIP-Ctx), has been shown to be able of utilizing the future context effectively, meanwhile with quite low model latency and computation cost. 

In this paper, we continue to improve mGRUIP-Ctx with two revisions: applying BN methods and enlarging model context. Experimental results on two Mandarin ASR tasks (8400 hours and 60K hours) show that, the revised mGRUIP-Ctx outperform LSTM with a large margin (11\% to 38\%). It even performs slightly better than a superior BLSTM on the 8400h task, with 33M less parameters and just 290ms model latency.

\end{abstract}

\section{Introduction}

It is typically shown to be beneficial for acoustic modeling to make full use of the future contextual information.  In the literature, there are variety of methods to realize this idea for different model architectures. For feed-forward neural network (FFNN), this context is usually provided by splicing a fixed set of future frames in the input representation\cite{DongYu}. The authors in \cite{FSMN1,FSMN2,FSMN3} proposed a model called feedforward sequential memory networks (FSMN), which is a standard FFNN equipped with some learnable memory blocks in the hidden layers to encode the long context information into a fixed-size representation. The time delay neural network (TDNN) \cite{SPL2,SPL17} is another FFNN architecture which has been shown to be effective in modeling long range dependencies through temporal convolution over context. 

As for unidirectional recurrent neural network (RNN), this is usually accomplished using a delayed prediction of the output labels\cite{ASRU6}. However, this method only provides quite limited modeling power of future context\cite{SPL}. While for bidirectional RNN, this is accomplished by processing the data in the backward direction using a separate RNN layer \cite{SPL4,SPL5,SPL6}. Although the bidirectional versions have been shown to outperform the unidirectional ones with a large margin \cite{SPL7,SPL8}, the latency of bidirectional models is significantly larger, making them unsuitable for online speech recognition. To overcome this limitation, chunk based training and decoding schemes such as context-sensitive-chunk (CSC) \cite{SPL10, Chen2016A} and latency-controlled (LC) BLSTM \cite{SPL7, SPL12} have been investigated. However, the model latency is still quite high, for example, the decoding latency of LC-BLSTM in \cite{SPL12} is about 600 ms. To overcome the shortcomings of the chunk-based methods, Peddinti \emph{et al.} \cite{SPL} proposed the use of temporal convolution, in the form of TDNN layers, for modeling the future temporal context while affording inference with frame-level increments. The proposed model is called TDNN-LSTM, and is designed by interleaving of temporal convolution (TDNN layers) with unidirectional long short-term memory (LSTM)  \cite{ASRU1,ASRU2,ASRU3,ASRU4} layers. This model was shown to outperform bidirectional LSTM in two automatic speech recognition (ASR) tasks, while enabling online decoding with a maximum latency of 200 ms \cite{SPL}. However, TDNN-LSTM's ability to model the future context comes from the TDNN part, whereas the LSTM itself is incapable of utilizing the future information effectively. 

Recently, in \cite{my}, we proposed a RNN model called minimal gated recurrent unit with input projection (mGRUIP), in which the inserted input projection forms a bottleneck and a context module called \emph{temporal convolution}, is specifically designed on it to model the future context. This model, mGRUIP with context module (mGRUIP-Ctx, for short in this paper), is able of utilizing the future context effectively and directly, meanwhile with quite low model latency and computation cost. 

In this work, we continue to improve the proposed model mGRUIP-Ctx. The revision is two-fold: First, we investigate how to use batch normalization (BN) \cite{BN-Google} in this model. 
The starting point is that we found the proposed mGRUIP-Ctx model performs not quite well if the training data contains a lot of noisy speech, for example, if the original clean data is perturbed with noise and reverberation. This prompts us to use batch normalization to improve the convergence of optimization process. In our previous work\cite{my}, batch normalization is only used on the cell ReLU activation to deal with numerical instabilities originating from the unbounded ReLU functions. In this work we find it is also beneficial to apply BN to the update gate as well. In the literature, batch normalization has been applied to RNN in different ways. In \cite{lightRNN49}, the authors suggest to apply it to input-to-hidden (ItoH) connections only, whereas in \cite{lightRNN50} the normalization step is extended to hidden-to-hidden (HtoH) transitions. Our finding in this work is slightly different with \cite{lightRNN49} and \cite{lightRNN50}. It is shown that the best method to apply BN in mGRUIP-Ctx is a \emph{hybrid} way, that is, applying BN to both ItoH and HtoH connections for the cell ReLU activation, while for the update gate, applying BN to ItoH only. Experimental results on several ASR tasks clearly demonstrate that doing so can speed up optimization and improve generalization, especially when the training data is augmented with perturbation.

The second revision is to enlarge the model context. In our previous work\cite{my}, the context module in mGRUIP-Ctx is restricted to just modeling the future context, leaving the history to be modeled by the RNN structure.  In this work, we release this restriction and allow the context module to model the future and history information simultaneously. It is empirically shown that this method is beneficial to the performance. Besides that, we also find that enlarging the order of future context can future improve the performance. It should be noted that this context extension brings quite limited additional parameters, thanks to the small dimensionality of the input projection.

With these two revisions, mGRUIP-Ctx's performance is improved significantly. On a 8400 hours Mandarin ASR task (1400 hours original data with 6-fold augmentation), the revised mGRUIP-Ctx provides 6\% to 10\% relative CER reduction over the previous one in \cite{my}, demonstrating the effectiveness of the revision methods. Compared to LSTM, the relative improvement is 18\% to 37\%, and the gain over TDNN-LSTM is about 5\% to 12\%. This model even outperforms a very strong baseline, BLSTM, with much less parameters and only 290 ms decoding latency. The proposed model's superiority is further verified on a much larger Mandarin ASR task, which contains 60K hours of speech in total (10K hours original data with 6-fold augmentation). On this task, the gain of mGRUIP-Ctx over LSTM and TDNN-LSTM is 11\% to 20\% and 3\% to 11\%, respectively. Compared with a much stronger baseline, TDNN-BLSTM, mGRUIP-Ctx just shows 2\% to 6\% relative loss, with advantages of much less parameters and much lower latency.

\section {Prerequisites}
\subsection {mGRU}
mGRU, short for minimal gated recurrent unit, is a revised version of GRU model. It is proposed by \cite{Bengio,LightRNN} and contains two modifications: removing the reset gate and replacing the hyperbolic tangent function with ReLU activation. The model is defined by the following equations (the layer index $l$ has been omitted for simplicity):
{\setlength\abovedisplayskip{3pt}
    \setlength\belowdisplayskip{3pt}
\begin{eqnarray}
  && \bm{\mathrm{z}}_t = \sigma(\bm{\mathrm{W}}_z\bm{\mathrm{x}}_t + \bm{\mathrm{U}}_z\bm{\mathrm{h}}_{t-1} + \bm{\mathrm{b}}_z) \\
  && \bm{\mathrm{\widetilde{h}}}_t = ReLU(BN(\bm{\mathrm{W}}_h\bm{\mathrm{x}}_t )+ BN(\bm{\mathrm{U}}_h\bm{\mathrm{h}}_{t-1})) \\
  && \bm{\mathrm{h}}_t = \bm{\mathrm{z}}_t*\bm{\mathrm{h}}_{t-1} + (1-\bm{\mathrm{z}}_t)*\bm{\mathrm{\widetilde{h}}}_t
\end{eqnarray}}
In particular, $\bm{\mathrm{z}}_t$ is a vector corresponding to the update gate, of which the activation is element-wise logistic sigmoid functions $\sigma(\cdot)$. $\bm{\mathrm{h}}_t$ represents the output state vector for the current frame $t$, and $\bm{\mathrm{\widetilde{h}}}_t$ is the candidate state obtained with a ReLU function. The parameters of the model are $\bm{\mathrm{W}}_z$, $\bm{\mathrm{W}}_h$ (the ItoH connections), $\bm{\mathrm{U}}_z$, $\bm{\mathrm{U}}_h$ (the HtoH weights), and the bias vector $\bm{\mathrm{b}}_z$. 

\subsection {mGRUIP}
mGRUIP is obtained by inserting a linear input projection layer into mGRU, leading to the following update equations:
{\setlength\abovedisplayskip{3pt}
    \setlength\belowdisplayskip{3pt}
\begin{eqnarray}
  && \bm{\mathrm{v}}_t = \bm{\mathrm{W}}_{v1}\bm{\mathrm{x}}_t + \bm{\mathrm{W}}_{v2}\bm{\mathrm{h}}_{t-1} \\
  && \bm{\mathrm{z}}_t = \sigma(\bm{\mathrm{W}}_z\bm{\mathrm{v}}_t + \bm{\mathrm{b}}_z) \\
  && \bm{\mathrm{\widetilde{h}}}_t = ReLU(BN(\bm{\mathrm{W}}_h\bm{\mathrm{v}}_t)) \\
  && \bm{\mathrm{h}}_t = \bm{\mathrm{z}}_t*\bm{\mathrm{h}}_{t-1} + (1-\bm{\mathrm{z}}_t)*\bm{\mathrm{\widetilde{h}}}_t 
\end{eqnarray}}
where the current input vector $\bm{\mathrm{x}}_t$ and the previous output state vector $\bm{\mathrm{h}}_{t-1}$, are compressed into a lower dimensional space by weight matrix $\bm{\mathrm{W}}_{v1}$ and $\bm{\mathrm{W}}_{v2}$ respectively and added together to get a projected vector $\bm{\mathrm{v}}_t$. Then the update gate activation $\bm{\mathrm{z}}_t$ and the candidate state vector $\bm{\mathrm{\widetilde{h}}}_t$ are calculated based on it.

\subsection {mGRUIP-Ctx}
The input projection forms a bottleneck in mGRUIP, on which we design a context module, \emph{temporal convolution}, to effectively model the future context\cite{my}. For the $l$ th layer ($l>1$), equation (4) now becomes:
{\setlength\abovedisplayskip{3pt}
    \setlength\belowdisplayskip{3pt}
\begin{eqnarray}
  && \bm{\mathrm{v}}_t^l = \bm{\mathrm{W}}_{v1}^l\bm{\mathrm{\widetilde{x}}}_t^l + \bm{\mathrm{W}}_{v2}^l\bm{\mathrm{h}}_{t-1}^l 
\end{eqnarray}
where $\bm{\mathrm{\widetilde{x}}}_t$ is the concatenation of the current input vector and the output state vector of preceding layer from serval future frames:
\begin{eqnarray}
  && \bm{\mathrm{\widetilde{x}}}_t = \left[\bm{\mathrm{x}}_t^l; \bm{\mathrm{h}}_{t+s\times i}^{l-1};\cdots;\bm{\mathrm{h}}_{t+s\times K}^{l-1}\right]
\end{eqnarray}}
In particular, $\bm{\mathrm{x}}_t^l$ is the input vector of layer $l$ ($\bm{\mathrm{x}}_t^l$ is actually $\bm{\mathrm{h}}_t^{l-1}$ since $l>1$), and $\bm{\mathrm{h}}_{t+s\times i}^{l-1}$ is the output state vector of layer $l-1$ on the $(t+s\times i)$th frame. $s\ge1$ is the step stride, $K\ge1$ is order of future context, and $1\le i \le K$ is loop index.

\section {Proposed Revisions}
\subsection {Applying Batch Normalization}
Our first refinement is to figure out the best way to apply BN in the proposed models. There are two possible locations to apply BN in the structure: the update gate and the cell activation. 
\subsubsection {BN for the update gate}
Three different BN methods for update gate are experimented: 
{\setlength\abovedisplayskip{3pt}
    \setlength\belowdisplayskip{1pt}
\begin{itemize}
  \setlength{\itemsep}{0pt}
  \setlength{\parsep}{0pt}
  \setlength{\parskip}{0pt}
  \item $\textbf{\emph{No BN}}$ \\
  Just as defined by equation (1) and (5) in Section 2 for model mGRU and mGRUIP(-Ctx), respectively.
  \item $\textbf{\emph{BN on ItoH only}}$ \\
  For mGRU, equation (1) now becomes:
  \begin{eqnarray}
    \bm{\mathrm{z}}_t = \sigma(BN(\bm{\mathrm{W}}_z\bm{\mathrm{x}}_t)+ \bm{\mathrm{U}}_z\bm{\mathrm{h}}_{t-1})
  \end {eqnarray}
  For mGRUIP(-Ctx), equation (5) now becomes:
  \begin {eqnarray}
    \bm{\mathrm{z}}_t = \sigma(BN(\bm{\mathrm{W}}_z\bm{\mathrm{v}}_{1t})+\bm{\mathrm{W}}_z\bm{\mathrm{v}}_{2t})
  \end{eqnarray}
  where $\bm{\mathrm{v}}_{1t}$ and $\bm{\mathrm{v}}_{2t}$ are the projected vector from $\bm{\mathrm{x}}_t$ (or $ \bm{\mathrm{\widetilde{x}}}_t $ for mGRUIP-Ctx) and $\bm{\mathrm{h}}_{t-1}$, respectively.
  \item $\textbf{\emph{BN on ItoH and HtoH}}$ \\
  For mGRU, equation (1) now becomes:
  \begin{eqnarray}
    \bm{\mathrm{z}}_t = \sigma(BN(\bm{\mathrm{W}}_z\bm{\mathrm{x}}_t)+ BN(\bm{\mathrm{U}}_z\bm{\mathrm{h}}_{t-1}))
  \end {eqnarray}
  For mGRUIP(-Ctx), equation (5) now becomes:
  \begin {eqnarray}
    \bm{\mathrm{z}}_t = \sigma(BN(\bm{\mathrm{W}}_z\bm{\mathrm{v}}_t))
  \end{eqnarray}
\end{itemize}}

\subsubsection {BN for the cell activation}
For the cell activation, \emph{No BN} can probably cause numerical instabilities for ReLU, thus only two methods are tried: 
{\setlength\abovedisplayskip{3pt}
    \setlength\belowdisplayskip{1pt}
\begin{itemize}
  \setlength{\itemsep}{0pt}
  \setlength{\parsep}{0pt}
  \setlength{\parskip}{0pt}
  \item $\textbf{\emph{BN on ItoH only}}$ \\
  We first try this method for mGRU model. Equation (2) now becomes:
  \begin{eqnarray}
    \bm{\mathrm{\widetilde{h}}}_t = ReLU(BN(\bm{\mathrm{W}}_h\bm{\mathrm{x}}_t )+ \bm{\mathrm{U}}_h\bm{\mathrm{h}}_{t-1})
  \end{eqnarray}
  It degrades the performance significantly (shown in Table 1), thus no further trial is needed for mGRUIP(-Ctx).
  \item $\textbf{\emph{BN on ItoH and HtoH}}$ \\
  Just as defined by equation (2) and (6) in Section 2 for model mGRU and mGRUIP(-Ctx), respectively.
\end{itemize}}

\subsection {Enlarging Model Context}
Our second revision is to enlarge the model context. The context module is used to model not only the future context but also the history information. Equation (9) now becomes:
{\setlength\abovedisplayskip{3pt}
    \setlength\belowdisplayskip{3pt}
\begin{eqnarray}
  \bm{\mathrm{\widetilde{x}}}_t &= \left[\bm{\mathrm{x}}_t^l; \bm{\mathrm{h}}_{t-s_1\times i}^{l-1};\cdots;\bm{\mathrm{h}}_{t-s_1\times K_1}^{l-1}; \right. \\
  \nonumber & \qquad \ \ \, \left. \bm{\mathrm{h}}_{t+s_2\times j}^{l-1};\cdots;\bm{\mathrm{h}}_{t+s_2\times K_2}^{l-1}\right]
\end{eqnarray}} 
where $s_1\ge1$ and $s_2\ge1$ is the step stride for the history and future context, respectively. $K_1\ge1$ is the order of history and $K_2\ge1$ is the order of future context.

{\setlength{\belowcaptionskip}{-0.2cm}
\setlength{\abovecaptionskip}{0.2cm}
\begin{table}[ht]\footnotesize
  \caption{Comparison of BN methods with mGRU and mGRUIP on Switchboard task.}
  \label{tab1}
  \centering
  \begin {tabular}{|cc|cc|ccc|}
  \toprule
   \multicolumn{2}{|c|} {Gate, BN on} & \multicolumn{2}{c|} {Cell, BN on} & \multicolumn{3}{c|}{WER (\%)} \\
   \midrule
   ItoH & HtoH & ItoH & HtoH & SWB & CHM & Total \\
   \midrule
   \multicolumn{7}{c}{mGRU} \\
  \midrule
   N & N & Y & Y & 10.2 & 20.6 & 15.5 \\
   Y & Y & Y & Y & \multicolumn{3}{c|}{diverge} \\
   Y & N & Y & Y & \textbf{10.1} & \textbf{19.9} & \textbf{15.1} \\
   Y & N & Y & N & 11.1 & 22.1 & 16.7 \\
  \midrule
  \multicolumn{7}{c}{mGRUIP} \\
  \midrule
  N & N & Y & Y & 9.8 & 19.0 & 14.5 \\
   Y & Y & Y & Y & 19.8 & 27.1 & 23.5 \\
   Y & N & Y & Y & \textbf{9.5} & \textbf{18.6} & \textbf{14.2} \\
  \bottomrule
  \end{tabular}
\end{table}}

\section {Experimental Settings}
This section will introduce the three ASR tasks used in this work. All the models in this paper are trained with LF-MMI objective function computed on 33Hz outputs \cite{SPL16}. 

\subsection {Switchboard ASR Task}
The training data is 309-hour Switchboard-I data. Evaluation is performed in terms of WER on the full Switchboard Hub5'00 test set, consisting of two subsets: Switchboard (SWB) and CallHome (CHE). The experimental setup follows \cite{SPL16}.WER results are reported after 4-gram LM rescoring of lattices generated using a trigram LM. Please refer to \cite{SPL16} for more details.

\subsection {Medium-Scale Mandarin ASR Task}
The second task is an internal medium-scale mandarin ASR task, which actually has two sub-tasks. The first one contains 1400 hours original mobile recording data. The second one is with 6 times data augmentation using speed perturbation (x3) \cite{Japan23} and noise/reverberation perturbation (x2) \cite{Japan24}, resulting in 8400 hours training data in total.

\subsection {Large-Scale Mandarin ASR Task}
The large-scale mandarin ASR task contains 10K hours original data and is augmented 6 times, resulting in 60K hours training data in total. The performance of mandarin ASR tasks are evaluated on five public-available test sets, including three clean and two noisy ones. The clean sets are:
\begin{itemize}
\setlength{\itemsep}{0pt}
\setlength{\parsep}{0pt}
\setlength{\parskip}{0pt}
\item AiShell\_dev and AiShell\_test: the development and test set of the released corpus AiShell-1\cite{Aishell}, containing 14326 and 7176 utterances, respectively.
\item THCHS-30\_Clean: the clean test set of THCHS-30 database\cite{thchs30}, containing 2496 utterances. 
\end{itemize}

The two noisy test sets are THCHS-30\_Car and THCHS-30\_Cafe, which are the corrupted version of THCHS-30\_Clean by car and cafeteria noise, respectively. The noise level is 0db. 

{
  \setlength{\abovecaptionskip}{0.1cm}
  \setlength{\belowcaptionskip}{-0.5cm}
\begin{figure} 
\flushleft
{\centering\includegraphics [width=8cm]{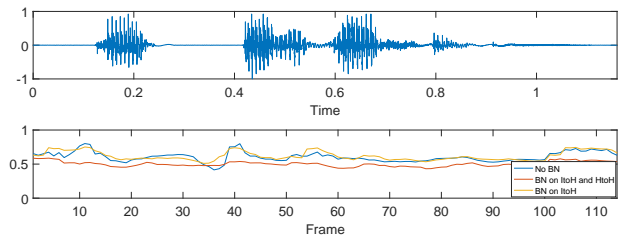}  
\caption{Average activation of update gate for three mGRUIP models trained with different BN methods.}  
\label{1}}  
\end{figure} }

\section {Experimental Results}
\subsection {Applying Batch Normalization}
\subsubsection {Switchboard ASR Task}
On this task, we compare different BN methods for two models: mGRU and mGRUIP, and the results are shown in Table 1. Both of them contain 5 layers, and each layer consists 1024 cells. The input projection layer for mGRUIP has 512 units.

Several observations can be obtained from Table 1. Firstly, comparing the first two lines of each model's results, we can see that applying BN to both ItoH and HtoH for update gate is quite harmful. The training of mGRU model even diverges. 
We think the reason is as below. The update gate is controlled by a sigmoid function, and is expected to learn to open or close at the right time. \emph{BN on ItoH and HtoH} will keep the input away from the saturated regime of this nonlinearity. This may be helpful when sigmoid serves as a hidden node activation. However, when it's used to control a gate, this will be harmful since it will make the gate half-closed or half-opened, which is opposite to the responsibility of the gate control. To verify this, 
we investigate the evolution of update activation when performing recognition. A speech segment is chosen from the test set (the utterance id is \emph{en\_4170-B\_064608-064704} with the text content \emph{That is quite a difference}), and the features are sent to the three mGRUIP models trained with different BN methods. The average activation of the update gate over cells (the 5th layer) for each frame is shown in Figure 1. It's very clear that applying BN to both ItoH and HtoH for the update gate will cause the gate activation fluctuating around 0.5, meaning the gate unable to effectively control the dependency between the history $\bm{\mathrm{h}}_{t-1}$ and candidate state $\bm{\mathrm{\widetilde{h}}}_t$. Another interesting finding is that, the average gate activations of \emph{No BN} and \emph{BN on ItoH} models are almost always greater than 0.5. We think it's because the speech signal is a sequence that evolves rather slowly, in which the past history can virtually always be helpful. 

{\setlength{\belowcaptionskip}{-0.2cm}
\setlength{\abovecaptionskip}{0.2cm}
\begin{table*}[h]\footnotesize
  \caption{Performance comparison of mGRUIP-Ctx with various settings of context module on 1400h Mandarin ASR task.}
  \label{tab3}
  \centering
  \begin {tabular}{c|c c c c |c|c|c|c|c|c}
  \toprule
   \multirow{3}*{Model} & \multicolumn {4} {c|} {\multirow{2}*{Layerwise Context Setting}} & \multirow{2}*{Latency} & \multicolumn {5} {c} {CER} \\
   \cline {7-11} 
   & & & & & & \multicolumn {2} {c|} {AiShell} & \multicolumn {3} {c} {THCHS-30} \\
   & 2 & 3 & 4 & 5 & (ms) & dev & test & Clean & Car & Cafe \\
   \midrule
   mGRUIP-Ctx-A & {0; $1 \times 1$} & {0; $1 \times 3$} & {0; $1 \times 3$} & {0; $1 \times 3$} & 170 & 4.61 & 5.56 & 10.34 & 10.67 & 41.89 \\
   mGRUIP-Ctx-B & {$1 \times 6$; $1 \times 1$} & {$1 \times 6$; $1 \times 3$} & {$1 \times 6$; $1 \times 3$} & {$1 \times 6$; $2 \times 3$} & 200 & 4.49 & \textbf{5.46} & 10.11 & 10.46 & 39.79 \\
   mGRUIP-Ctx-C & {$2 \times 6$; $1 \times 1$} & {$2 \times 6$; $1 \times 3$} & {$2 \times 6$; $1 \times 3$} & {$2 \times 6$; $2 \times 3$} & 200 & 4.49 & 5.50 & 10.08 & 10.45 & \textbf{38.41} \\
   mGRUIP-Ctx-D & {$1 \times 6$; $1 \times 1$} & {$1 \times 6$; $1 \times 3$} & {$1 \times 6$; $1 \times 6$} & {$1 \times 6$; $2 \times 6$} & 290 & \textbf{4.47} & 5.47 & \textbf{9.95} & \textbf{10.36} & 38.76 \\
  \bottomrule
  \end{tabular}
\end{table*}}

The second observation from Table 1 is that we should apply BN to both ItoH and HtoH for the cell ReLU activation (comparing the last two lines of mGRU's results). It's possibly because applying BN this way will provide the best numerical stabilities for ReLU. Finally, we can conclude from Table \ref{tab1} that, the best method to apply BN in mGRU related models is a hybrid way, that is, applying BN to just ItoH for update gate, and for cell ReLU activation, applying BN to both ItoH and HtoH.

{\setlength{\belowcaptionskip}{-0.2cm}
  \setlength{\abovecaptionskip}{0.2cm}
  \begin{table}[ht]\footnotesize
  \caption{Performance of different models on Medium-Scale Mandarin ASR task.}
  \label{tab2}
  \centering
  \begin {tabular}{|c|c|c|c|c|c|c|}
  \toprule
   \multirow{3}*{Model} & \multirow{2}*{Gate} & \multicolumn {5} {c|} {CER} \\
   \cline {3-7} 
   & & \multicolumn {2} {c|} {AiShell} & \multicolumn {3} {c|} {THCHS-30} \\
   & BN & dev & test & Clean & Car & Cafe \\
   \midrule
   \multicolumn{7}{c}{1400 Hours Original Data} \\
   \midrule
   TDNN-LSTM                          & -          & 4.81 & 5.98 & 10.97 & 11.38 & 44.20 \\
   \cmidrule {1-2} \cmidrule {3-7}
   \multirow{2}*{mGRUIP-Ctx}& No\cite{my}      & 4.66 & 5.71 & 10.38 & 10.77 & \textbf{40.26} \\
                                                    & ItoH   &  \textbf{4.61} & \textbf{5.56} & \textbf{10.34} & \textbf{10.67} & 41.89 \\
   \midrule 
   \multicolumn{7}{c}{8400 Hours Augmented Data} \\
   \midrule
   TDNN-LSTM                          & -          & 4.50 & 5.42 & 10.11 & 10.30 & 23.86 \\
   \cmidrule {1-2}\cmidrule {3-7}
   \multirow{2}*{mGRUIP-Ctx}& No      & 4.45 & 5.42 & 10.24 & 10.34 & 23.43 \\
                                                    & ItoH   &  \textbf{4.33} & \textbf{5.26} & \textbf{10.00} & \textbf{10.02} & \textbf{22.35} \\
 \bottomrule
  \end{tabular}
\end{table}}

\subsubsection {Medium-Scale Mandarin ASR Task}
According to the findings in Section 5.1.1, we apply the optimal BN methods to the model mGRUIP-Ctx on the medium-scale mandarin ASR task, and the results are shown in Table \ref{tab2}. The baseline model is TDNN-LSTM \cite{SPL} which has interleaving 7 TDNN and 3 LSTM layers. The model mGRUIP-Ctx contains 5 hidden layers, each one has 2560 cells and a 256-dimensional input projection layer. The settings of context order $K$ and step stride $s$ for each layer is same as our previous work \cite{my}.

According to Table \ref{tab2}, with 1400 hours original training data, mGRUIP-Ctx with no BN on update gate gives 3\% to 5\% relative gain over TDNN-LSTM. However, this gain disappears after the training data is augmented with perturbation. It can be attributed to the different difficulties of these two sub-tasks, of which model learning with perturbed data is harder than with original data. After BN is correctly applied, mGRUIP-Ctx performs better than TDNN-LSTM on the harder sub-task, demonstrating the effectiveness of the BN method. Moreover, the training curve reveals that the optimization process becomes much stable (not shown in this paper).

\subsection {Enlarging Model Context}
\subsubsection {Medium-Scale Mandarin ASR Task}
Our second revising method is to enlarge the model context. We compare the performance of mGRUIP-Ctx with different settings of context order $K_1$, $K_2$ and step stride $s_1$, $s_2$ (BN methods are correctly used). For fast experiments, all the models are trained with 1400 hours original data, and the results are reported in Table \ref{tab3}. The setting of context order and stride for each layer is represented with a format of $\{K_1\times s_1$;$K_2 \times s_2\}$.

From Table \ref{tab3}, comparing  mGRUIP-Ctx-A and  mGRUIP-Ctx-B, it can be seen that allowing context module to model the history information as well is quite beneficial to the performance (mGRUIP-Ctx-A is just the model used in Section 5.1.2). In addition, increasing the order of future context can further improve the performance, with a cost of additional model latency ( mGRUIP-Ctx-D vs.  mGRUIP-Ctx-B). 

Next, the best model mGRUIP-Ctx-D is compared with several baselines, including BLSTM, TDNN-LSTM and LSTM, on the 8400 hours data-augmented task, and the results are presented in Table 4. BLSTM and LSTM both contain 5 hidden layers. Each layer of LSTM and each directional sub-layer of BLSTM, have 1024 cells and 512 linear projection units. TDNN-LSTM is the model used in Section 5.1.2.

{\setlength{\belowcaptionskip}{-0.2cm}
\setlength{\abovecaptionskip}{0.2cm}
\begin{table}[ht]\footnotesize
  \caption{Performance of different models on 8400h Mandarin ASR task.}
  \label{tab4}
  \centering
  \begin {tabular}{|c|c|c|c|c|c|c|}
  \toprule
   \multirow{3}*{Model} & \multirow{2}*{\scriptsize{Latency}} & \multicolumn {5} {c|} {CER} \\
   \cline {3-7} 
   & & \multicolumn {2} {c|} {AiShell} & \multicolumn {3} {c|} {THCHS-30} \\
   & (ms) & dev & test & Clean & Car & Cafe \\
   \midrule
   BLSTM           & 2020 & 4.24 & 5.05 & 9.79 & 9.87 & 22.82 \\
   TDNN-LSTM & 210   & 4.50 & 5.42 & 10.11 & 10.3 & 23.86 \\
   LSTM             & 70     &  5.07 & 6.23 & 12.06 & 11.82 & 33.67 \\
   \midrule 
   \scriptsize{mGRUIP-Ctx-D}  & 290 & \textbf{4.13} & \textbf{5.03} & \textbf{9.56} & \textbf{9.71} & \textbf{20.94} \\
   \midrule
   CERR (\%) &   -   & 18.5 & 19.3              & 20.7 & 17.9 & 37.8 \\
 \bottomrule
  \end{tabular}
\end{table}}

The \emph{CERR} line in Table 4 is the relative gain of mGRUIP-Ctx-D over LSTM, ranging from 18\% to 38\% for different test sets, indicating that the proposed model performs significantly better than LSTM. Compared with TDNN-LSTM, mGRUIP-Ctx-D gives 5\% to 12\% relative CER reduction with 12.4M less parameters (22.4M vs. 34.8M) and 80 ms additional latency (290ms vs. 210ms). mGRUIP-Ctx-D even outperforms BLSTM with much lower model latency (290ms vs. 2020ms) and much less model parameters (22.4M vs. 55.3M). Compared to the performance of mGRUIP-Ctx before revision (the next-to-last line of Table \ref{tab2}), mGRUIP-Ctx-D shows 6\% to 10\% relative gain, which demonstrates the effectiveness of the two proposed revising methods. 

\subsubsection {Large-Scale Mandarin ASR Task}
Finally, mGRUIP-Ctx-D's superiority is verified on a much larger mandarin ASR task, which contains 60K hours speech data. We train a TDNN-BLSTM model as one baseline, which is stronger than BLSTM and contains 3 TDNN layers and 5 BLSTM layers. All of the other models have the same structure as Section 5.2.1. Results are shown in Table 5.

{\setlength{\belowcaptionskip}{-0.2cm}
 \setlength{\abovecaptionskip}{0.2cm}
\begin{table}[ht]\footnotesize
  \caption{Performance of different models on 60K hour Mandarin ASR task.}
  \label{tab5}
  \centering
  \begin {tabular}{|c|c|c|c|c|c|}
  \toprule
   \multirow{3}*{Model} &  \multicolumn {5} {c|} {CER} \\
   \cline {2-6} 
   & \multicolumn {2} {c|} {AiShell} & \multicolumn {3} {c|} {THCHS-30} \\
   &  dev & test & Clean & Car & Cafe \\
   \midrule
   TDNN-BLSTM    & \textbf{3.55} & \textbf{4.21} & \textbf{8.72} & \textbf{8.85} & \textbf{18.73} \\
   TDNN-LSTM      & 3.90 & 4.68 & 9.55 & 9.65 & 21.53 \\
   LSTM                  &  4.32 & 5.26 & 10.3 & 10.33 & 23.84 \\
   \midrule 
   mGRUIP-Ctx-D   & 3.71 & 4.52 & 9.09 & 9.16 & 19.06 \\
   \midrule
   CERR (\%)    & 14.1 & 14.1 & 11.8 & 11.3 & 20.1 \\
 \bottomrule
  \end{tabular}
\end{table}}

According to Table 5,  the relative improvement of mGRUIP-Ctx-D over LSTM is 11\% to 20\%, and the gain over TDNN-LSTM is 3\% to 11\%. mGRUIP-Ctx-D performs slightly worse (2\% to 6\% relative) than the strong baseline, TDNN-BLSTM, but has two advantages: much less parameters and much lower latency.

\section{Conclusions}
In this paper, we improve our previous proposed model mGRUIP-Ctx with two revisions: applying BN methods and enlarging model context. After revising, mGRUIP-Ctx outperform LSTM with a large margin. It even performs slightly better than a superior BLSTM on one task, with much less parameters and much lower latency. 

\bibliographystyle{IEEEtran}

\bibliography{mybib}


\end{document}